\documentclass[lettersize,journal]{IEEEtran}
\usepackage{amsmath,amsfonts}
\usepackage{algorithm}
\usepackage{array}
\usepackage[caption=false,font=normalsize,labelfont=sf,textfont=sf]{subfig}
\usepackage{textcomp}
\usepackage{stfloats}
\usepackage{url}
\usepackage{verbatim}
\usepackage{graphicx}
\usepackage{cite}
\usepackage{booktabs}
\usepackage{siunitx}
\usepackage{caption}
\usepackage{algpseudocode} 
\usepackage{multirow}
\usepackage{adjustbox}
\usepackage{cellspace}
\usepackage{tabularx}
\usepackage{enumitem}
\usepackage{soul}

\newcommand\highlightsize{.035}
\newcommand\heatmapsize{0.33}

\newcolumntype{D}{>{\hfill}N{3}{2}<{\hfill}}

\newcommand\textconfidence{\fontsize{6pt}{6pt}\selectfont}

\usepackage{authblk}

\begin{document}

\title{Adversarial sample generation and training using geometric masks for accurate and resilient license plate character recognition}

\author[1]{\IEEEauthorblockN{Bishal Shrestha\textsuperscript{1,}\IEEEauthorrefmark{1},
Griwan Khakurel\textsuperscript{2,}\IEEEauthorrefmark{2},
Kritika Simkhada\textsuperscript{2,}\IEEEauthorrefmark{3}, 
Badri Adhikari\textsuperscript{3,}\IEEEauthorrefmark{4}}\\
\IEEEauthorblockA{\textsuperscript{1}Department of Computer Science, University of Miami, Coral Gables, Florida, USA}\\
\IEEEauthorblockA{\textsuperscript{2}Advanced College of Engineering and Management, Tribhuvan University, Nepal}\\
\IEEEauthorblockA{\textsuperscript{3}Department of Computer Science, University of Missouri-St. Louis, St. Louis, USA}\\
\IEEEauthorrefmark{1}bishalshrestha@miami.edu,
\IEEEauthorrefmark{2}grkhakurel@gmail.com,
\IEEEauthorrefmark{3}simkhadakritika12@gmail.com,
\IEEEauthorrefmark{4}adhikarib@umsl.edu
% <-this % stops an unwanted space
}

\maketitle

\begin{abstract}
\normalfont{\textbf{Background.} Reading dirty license plates accurately in moving vehicles is challenging for automatic license plate recognition systems. Moreover, license plates are often intentionally tampered with a malicious intent to avoid police apprehension. Usually, such groups and individuals know how to fool the existing recognition systems by making minor unnoticeable plate changes. Designing and developing deep learning methods resilient to such real-world `attack' practices remains an active research problem. As a solution, this work develops a resilient method to recognize license plate characters.
\textbf{Methods.} Extracting 1057 character images from 160 Nepalese vehicles, as the first step, we trained several standard deep convolutional neural networks to obtain 99.5\% character classification accuracy. On adversarial images generated to simulate malicious tampering, however, our model's accuracy dropped to 25\%. Next, we enriched our dataset by generating and adding geometrically masked images, retrained our models, and investigated the models' predictions.
\textbf{Results.} The proposed approach of training with generated adversarial images helped our adversarial attack-aware license plate character recognition (AA-LPCR) model achieves an accuracy of 99.7\%. This near-perfect accuracy demonstrates that the proposed idea of random geometric masking is highly effective for improving the accuracy of license plate recognition models. Furthermore, by performing interpretability studies to understand why our models work, we identify and highlight attack-prone regions in the input character images.
In sum, although Nepal's embossed license plate detection systems are vulnerable to malicious attacks, our findings suggest that these systems can be upgraded to close to 100\% resilience.}
\end{abstract}


\section{Introduction}

\IEEEPARstart{F}{or} effective e-governance, automatic license plate recognition systems should be accurate and reliable. Unfortunately, these deep neural network-based systems are often vulnerable and under attack. Recent studies show that deep neural networks are vulnerable to adversarial attacks, where subtle, imperceptible, or malicious modifications to input data can lead to misleading and erroneous predictions \cite{Akhtar_Mian_Kardan_Shah_2021}. Real-world AI applications can be at a loss due to these adversarial attacks \cite{eykholt2018robust,wang2021research,fang2023state}. A successful adversarial attack could have disastrous consequences, potentially allowing criminals to evade identification, compromising national security, and undermining the effectiveness of law enforcement efforts. In the case of automatic License Plate Character Recognition (LPCR) systems, tiny manipulations can be maliciously employed in each character of the license plates, allowing vehicles to evade police apprehension or engage in criminal activities undetected. Unfortunately, despite the high volume of theoretical research on adversarial attack algorithms and defense mechanisms against these attacks \cite{khamaiseh2022adversarial, wang2021adversarialagain}, studies that focus on practical real-life attack probabilities are missing.

Using a small license character dataset obtained from vehicles in Nepal, in this research, we elucidate how existing deep-learning algorithms for license plate character recognition can be easily improved against adversarial attacks. Specifically, we first discuss a novel adversarial attack algorithm to generate effective adversarial license plate character images. These adversarial images are similar to the images that may be observed in real-life settings. We then demonstrate how existing algorithms and methods are highly vulnerable to such adversarial samples. Next, we demonstrate that a deep learning model can be made considerably robust to such adversarial attacks, by simply training with the adversarial samples. Finally, we present several findings of our interpretability experiments to understand when these methods are inaccurate. 

The paper is structured as follows: Section 2 investigates related works in this domain, Section 3 outlines the methodology, Section 4 presents and analyzes the results of the experiment, and Section 5 concludes the paper with a discussion of future work.

% === II. Related Works ========================y
% =================================================================================
\section{Related Works}
Every year, neural networks have brought outstanding advancements in the License Plate Recognition (LPR) system \cite{anagnostopoulos2008license, anagnostopoulos2006license, kocer2011artificial, li2018toward}. The automatic license plate recognition system proposed by Chang et al. \cite{ALPR} consists of two modules: a license plate locating module and a license number identification module. In Nepal, the government has introduced embossed license plates replacing the traditional handwritten ones to implement a robust LPR system. Interestingly, recent studies have shown that adversarial examples expose vulnerabilities in the architectures of deep neural network \cite{szegedy2013intriguing} as security of machine learning \cite{barreno2010security} has been of concern for a long time. A survey on the threat of adversarial attacks on Deep Learning in Computer Vision shows the transferability of such attacks on real-life scenarios \cite{akhtar2018threat}. 

Goodfellow et al. \cite{goodfellow2014explaining} proposed FGSM (Fast Gradient Sign Method), which is to calculate one-step gradient update along the direction of the sign of gradient at each pixel. This method provided a meager computational cost. Since then, researchers have proposed different techniques to generate adversarial examples. Moosavi-Deezfooli et al. \cite{moosavi2016deepfool} proposed Deepfool to compute a minimal norm of adversarial perturbation for a given image in an iterative manner which provided less perturbation compared to FGSM. Carlini and Wagner \cite{carlini2017towards} proposed the C\&W method, which defeated the defensive distillation for targeted networks. Shu and Vargas et al. \cite{shu2020adversarial} proposed a single-pixel attack method using the genetic algorithm to deceive a neural network by altering a single pixel. These adversarial attacks can be used to create adversarial samples for an LPR system that is not perceived by humans and can only be detected by the system \cite{kwon2021adv}. However, these attacks cannot be simulated easily regarding an actual license plate. Quian et al. \cite{qian2020spot} proposed a spot evasion attack method for the LPR system, which uses a genetic algorithm to find the optimal perturbation position. This simulated a more real-world scenario that could be used to tamper with the license plate. However, this approach only looked into spots that disregard other shapes of different sizes, which can be easier to reproduce in real life. 

Studies by Kurakin et al. \cite{Kurakin2016AdversarialEI} showcase the ease with which adversarial examples can mislead DNN classifiers, even when modifications are undetectable to human observers. Furthermore, the introduction of physically feasible adversarial stickers highlights the significance of investigating stealthy attack methods that exploit real-world objects, reinforcing the importance of resilience in the face of physical adversarial threats \cite{Wei2021AdversarialSA}. In the realm of real-world object detection, the universal background adversarial attack method proposed by Yidan Xu et al. \cite{Xu2022UniversalPA} modifies local pixel information around target objects using a single background image, underscoring the relevance of studying robustness against physical attacks. Moreover, leveraging the natural phenomenon of shadows, \cite{Zhong2022ShadowsCB} presents a stealthy and effective physical-world adversarial attack that generates inconspicuous adversarial examples, further highlighting the necessity for resilience against emerging attack methods utilizing natural phenomena. Recent studies have also highlighted the vulnerability of deep neural networks (DNNs) to physical-world attacks using light beams. The Adversarial Laser Beam (AdvLB) attack \cite{Duan2021AdversarialLB} and the Adversarial Laser Spot (AdvLS) attack \cite{Hu2022AdversarialLS} demonstrate the efficacy of manipulating laser beam parameters to deceive DNNs, emphasizing the need for resilient models against light-based adversarial threats.

On the basis of threat, Papernot et al. \cite{papernot2017practical} presented effective results on the practical black-box attack against machine learning that is capable of evading defense strategies previously found to generate adversarial examples harder. Previous work of Szegedy et al. \cite{yosinski2014transferable} showed that adversarial samples have the property of transferability, which shows that such adversarial samples can be misclassified across models.

In the domain of defending deep neural networks from physically realizable attacks, a study \cite{Wu2019DefendingAP} demonstrates the limited effectiveness of existing robust model learning methods against prominent physical attacks on image classification. In response, they propose a new abstract adversarial model, rectangular occlusion attacks, and leverage adversarial training with this model to achieve high robustness against physically realizable attacks. Similarly, another study \cite{Kim2022DefendingPA} focuses on defending object detection systems against adversarial patch attacks by introducing a defense method based on ``Adversarial Patch-Feature Energy" (APE). The APE-based defense exhibits impressive robustness against adversarial patches in both digital and physical settings, providing a promising approach for countering physical adversarial attacks in critical systems like autonomous vehicles.

% === III. The Proposed Method =======================================
% =================================================================================
\section{Methodology}

The methodology employed in this research study outlines the step-by-step process followed to investigate and analyze the robustness of a license plate character recognition model against proposed adversarial attack. The methodology encompasses various stages, including image acquisition, preprocessing, character segmentation and labeling, model development, adversarial attack implementation, external validation of adversarial samples, and adversarial training. Each step was carefully executed to ensure the reliability and validity of the experimental results. The following sections provide an overview of the steps in our overall experiment of this study.

\textbf{Step 1: Image Acquisition.} High-quality license plate images were collected from the VFTC government office, which is responsible for license plate distribution. The images were acquired using a SONY A6000 Digital camera. Both front and rear license plate images were included.

\textbf{Step 2: Image Preprocessing.} The acquired images were filtered for noise, the resolution was scaled down, and license plate localization was performed to remove unnecessary backgrounds.

\textbf{Step 3: Character Segmentation and Labeling.} The characters were segmented from the preprocessed images using a fixed mask, and they were labeled accordingly.

\textbf{Step 4: Character Recognition Model Development.} A highly accurate multiclass CNN-based classification model was trained using the clean and labeled dataset.

\textbf{Step 5: Adversarial Attack Implementation.} The proposed exhaustive geometric mask-based adversarial algorithm was employed to generate adversarial samples, targeting the benign images within the multiclass CNN-based classification model.

\textbf{Step 6: External Validation of Adversarial Samples.} The generalizability of the generated adversarial samples in deceiving character recognition models beyond their original target was confirmed through evaluation with an external character recognition model.

\textbf{Step 7: Adversarial Training.} The multiclass CNN-based character classification model, was retrained, incorporating the randomly generated adversarial samples using the proposed attack method into the training dataset.

\subsection{Dataset}
For this study, three different types of datasets have been used. The first type is the I-1057 dataset, which consists of the pre-processed images of the collected data. The second type is the I-Hard-1057 dataset, which consists of adversarial images generated using the Exhaustive Geometric patch-based adversarial attack on the License Plate Character Recognition (LPCR) model. The third type is the I-Adversarial-Train dataset, which consists of equally distributed I-1057 images and randomly generated adversarial samples using geometric patches (horizontal, vertical, and circular), which is used for adversarial training of LPCR model to generate Adversarial Attack-aware License Plate Character Recognition (AA-LPCR) model.

\begin{figure}[!t]
\centering
\subfloat[]{\includegraphics[width=0.48\linewidth]{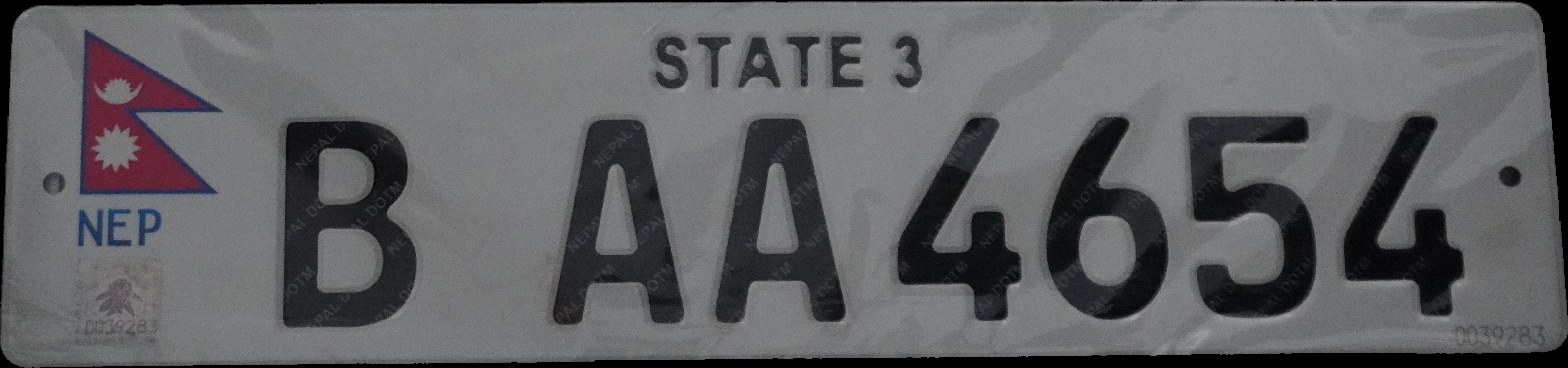}%
\label{fig_private_front}}
\hfil
\subfloat[]{\includegraphics[width=0.48\linewidth]{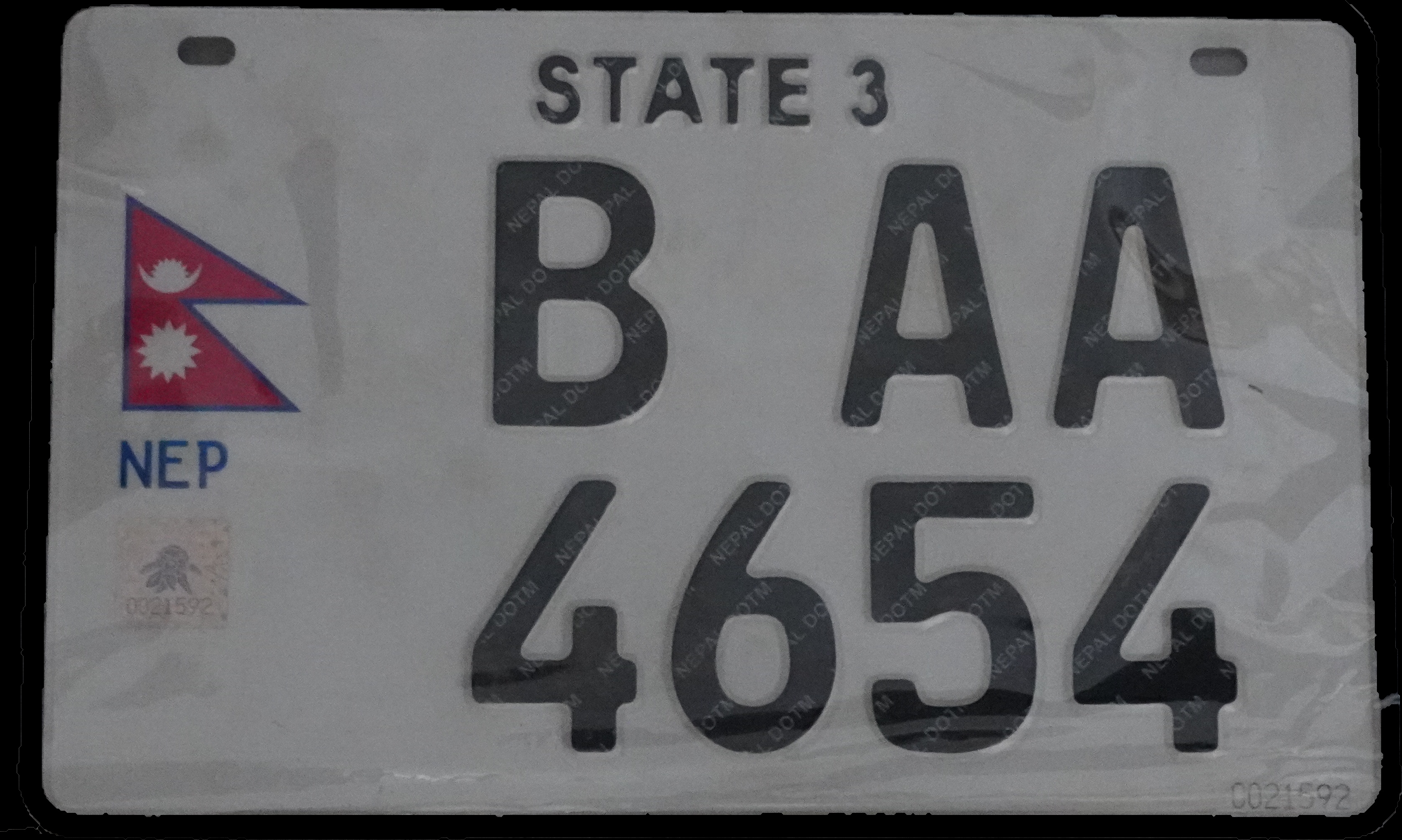}%
\label{fig_private_back}}
\hfil
\subfloat[]{\includegraphics[width=0.48\linewidth]{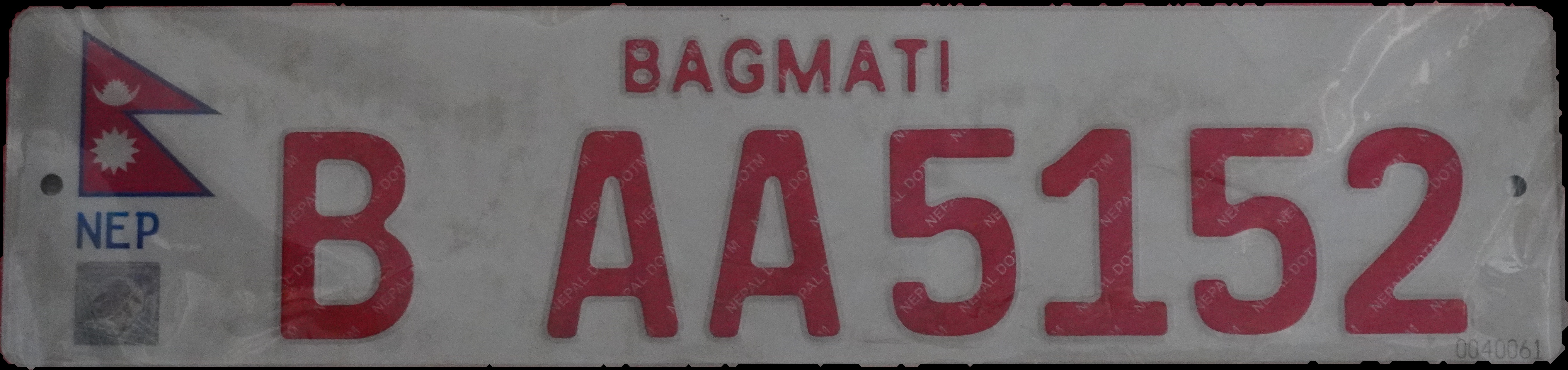}%
\label{fig_governmental_front}}
\hfil
\subfloat[]{\includegraphics[width=0.48\linewidth]{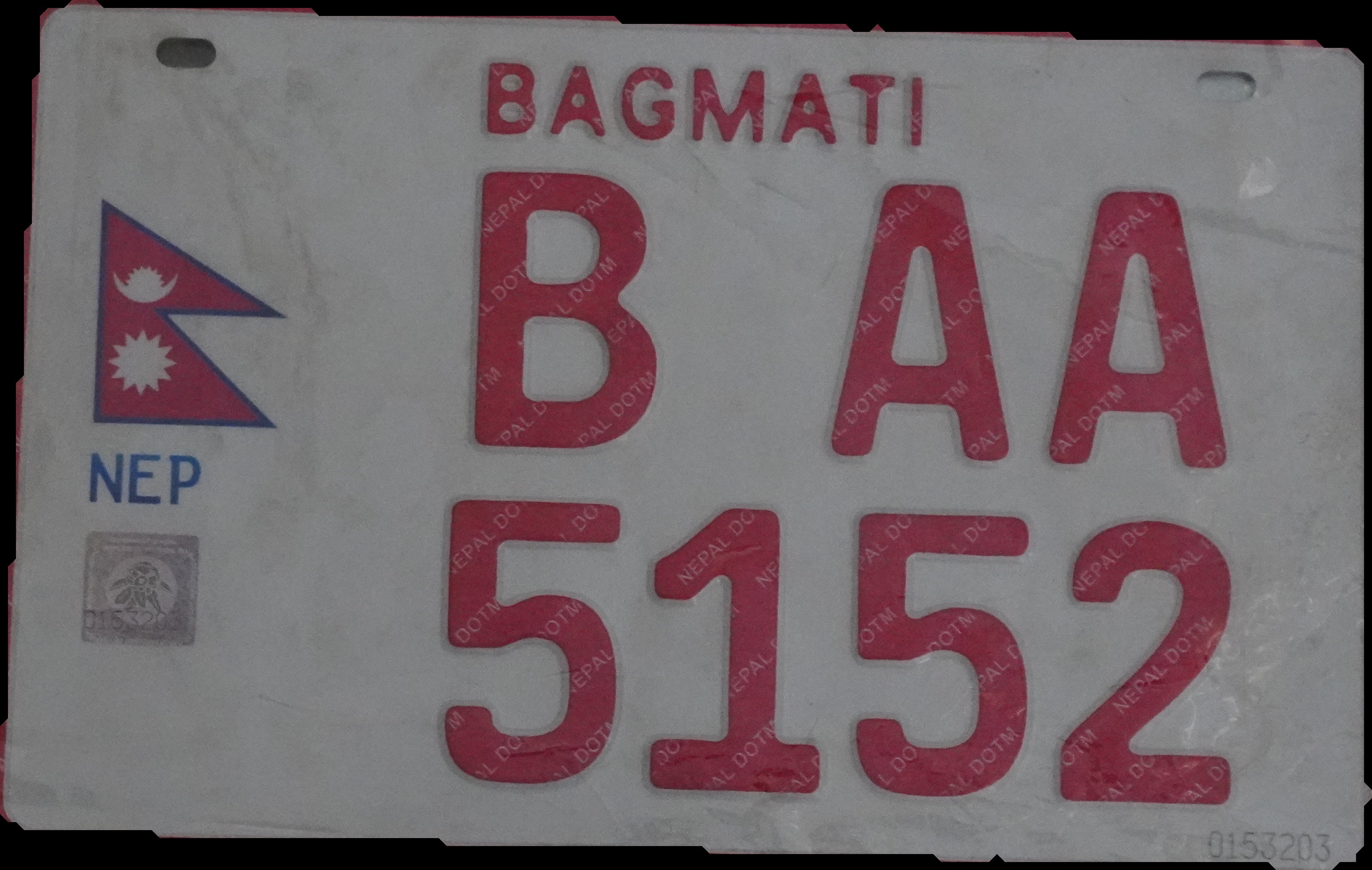}%
\label{fig_governmental_back}}
\hfil
\subfloat[]{\includegraphics[width=0.48\linewidth]{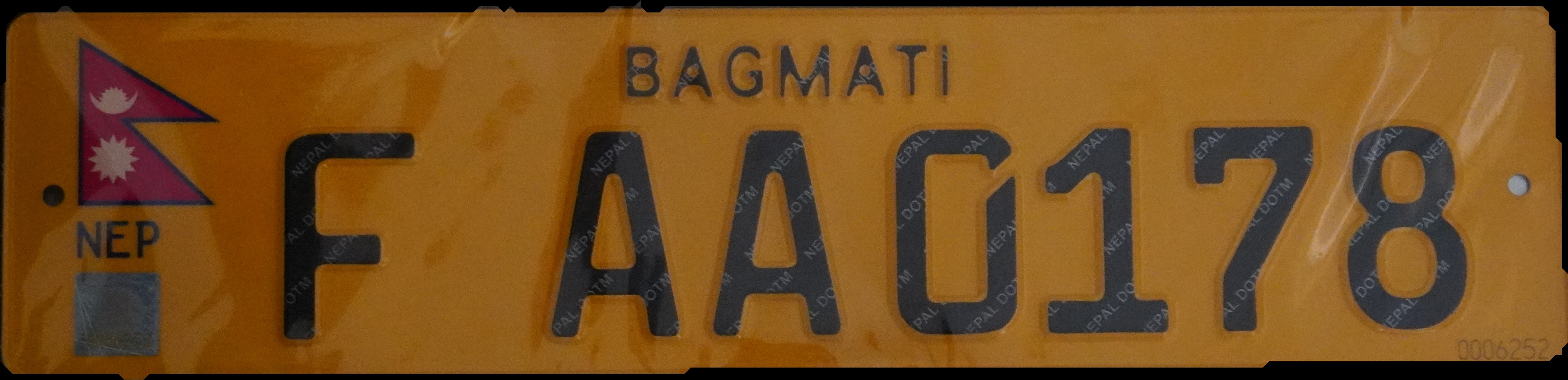}%
\label{fig_taxi_front}}
\hfil
\subfloat[]{\includegraphics[width=0.48\linewidth]{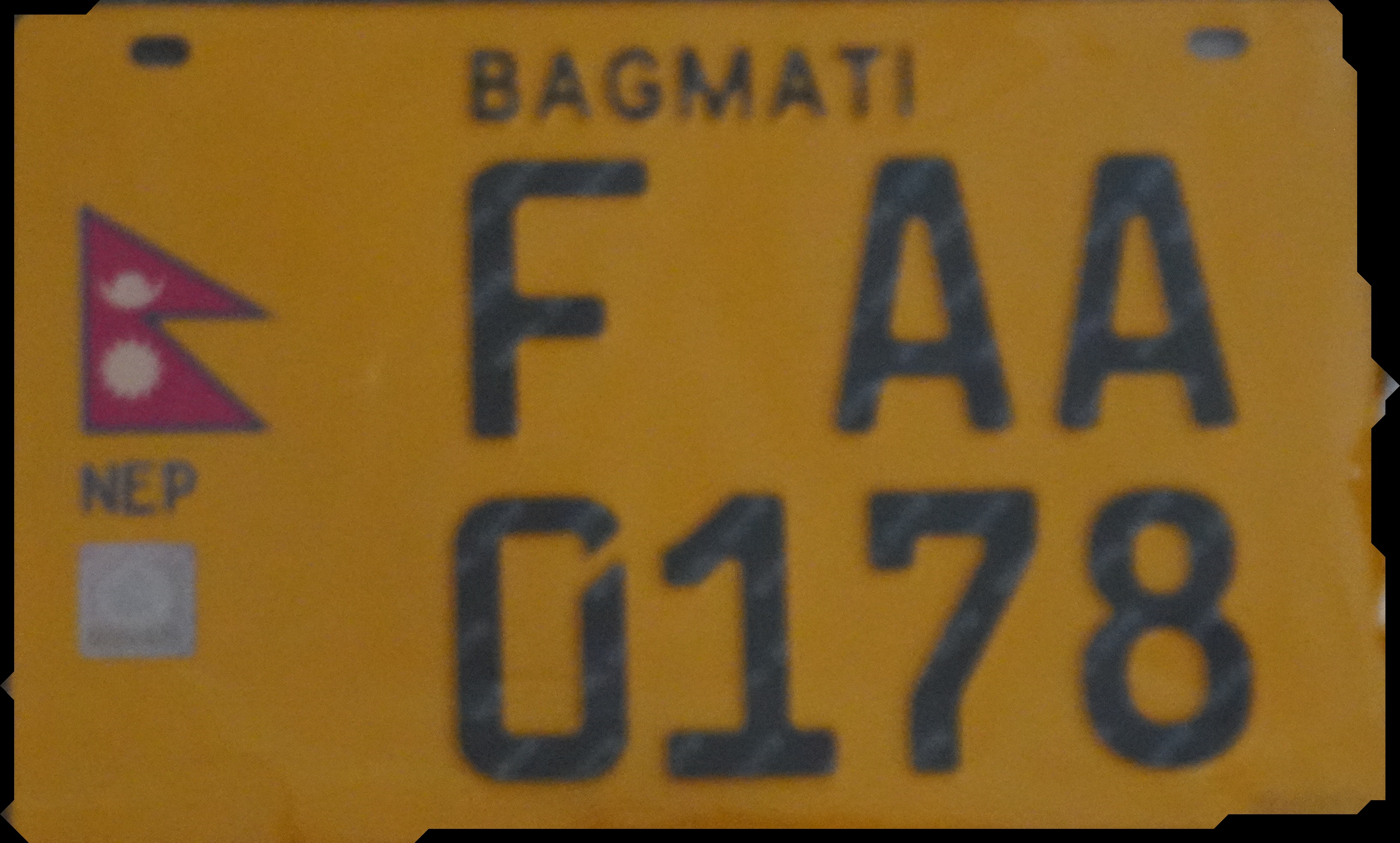}%
\label{fig_taxi_back}}

\caption{Example license plate images from captures from the front (first column) and rear (second column) of vehicles. (a) and (b) are from private vehicles, (c) and (d) are from governmental vehicles, and (e) and (f) are from public vehicles.} 
\label{fig_license_plate}
\end{figure}

The images of Nepal's Embossed Licence Plates that had been introduced to replace existing hand-written license plates were required, as shown in Fig \ref{fig_license_plate}. These data were acquired from the Vehicle Fitness Testing Centre (VFTC), which consisted of License Plates of four-wheeler vehicles, both front and rear, from State 3 for Private, Public, and Governmental vehicles. A total of 160 samples were collected using SONY A6000 Digital camera. The collected samples were filtered manually for noisy images. The collected images were of high resolution. To simulate a natural License Plate Recognition (LPR) system, the images were scaled down, followed by License Plate Localization using OpenCV.

For the segmentation of characters, a fixed mask was used by generalizing the position of the characters. The characters from License Plate were extracted using the mask, which was later labeled, giving us the I-1057 dataset. This I-1057 dataset consisted of 1057 images of characters segmented from Nepal's Embossed License Plate with the dimensions of 105x160x3 and consisted of characters 0, 1, 2, 3, 4, 5, 6, 7, 8, 9, A, B, and F. The I-1057 dataset was subjected to augmentation (Rotation and Blur) to imitate real-life scenarios during the model's training. 

\subsection{CNN-based License Plate Character Recognition (LPCR) model}
With the recognition of a CNN-based classifier, LPR systems are built with a state-of-art approach. For this study, a CNN-based classification model is used for better recognition of characters on the license plate. The CNN architecture was designed from scratch, achieving optimum time and space complexity, and the model was trained using the I-1057 dataset with Gaussian blur and rotation as augmentation using Pytorch. This augmented I-1057 dataset was subjected to a train-validation split of 80-20, which was fed into the model while keeping the image dimension of 105x160x3 (no grey scaling was done). 

On training the LPCR model, the optimum parameters were found to be: Learning Rate of 0.001, Momentum of 0.9, Batch size of 64, and Epochs of 80. The architecture of the CNN model can be seen in \textbf{Table \ref{tab_CNN_architecture}}.

After the training process, 10-fold cross-validation was done to validate the model's ability to get the most accurate prediction of embossed characters.

\begin{table}[ht]
\centering
\caption{CNN architecture. The number inside the bracket represents the number of neurons on the connected layers. On the convolution layer, the numbers inside the bracket represent the height, width, and number of filters, respectively.}
\begin{tabular}[t]{>{\raggedright}p{0.7\linewidth}>{\centering\arraybackslash}p{0.2\linewidth}}
\toprule
Layer Type  &   Shape\\
\midrule
Convolution + BatchNorm + ReLU      &   [3, 3, 16]\\
Convolution + BatchNorm + ReLU      &   [3, 3, 16]\\
Max pooling                         &   [2]\\
Convolution + BatchNorm + ReLU      &   [3, 3, 32]\\
Convolution + BatchNorm + ReLU      &   [3, 3, 32]\\
Max pooling                         &   [2]\\
Convolution + BatchNorm + ReLU      &   [3, 3, 64]\\
Convolution + BatchNorm + ReLU      &   [3, 3, 64]\\
Max pooling                         &   [2]\\
Convolution + BatchNorm + ReLU      &   [3, 3, 128]\\
Convolution + BatchNorm + ReLU      &   [3, 3, 128]\\
Convolution + BatchNorm + ReLU      &   [3, 3, 128]\\
Max pooling                         &   [2]\\
Fully connected + ReLU              &   [2304]\\
Dropout                             &   [0.5]\\
Fully connected + ReLU              &   [500]\\
Softmax                             &   [13]\\
\bottomrule
\end{tabular}
\label{tab_CNN_architecture}
\end{table}%

\subsection{Adversarial Attack}
% [To be removed: What is adversarial attack]
Adversarial attacks use an adversarial image to fool the neural networks by providing wrong information regarding the original image with the intention that the neural network misclassifies the original image \cite{szegedy2013intriguing}.

Popular algorithms such as FGSM, C\&W, and DeepFool are commonly used to generate adversarial examples to attack or test a model. Depending upon the domain and purpose of the attack, the usability of such an attack varies accordingly. Despite producing optimum adversarial examples, such gradient-based attacks are difficult to reproduce in Smart Embossed License Plates. As shown in figure \ref{fig:heatmap}, a calculated pixelated noise can be seen on the generated adversarial examples. This pixelated noise is difficult to be replicated in a real-life scenario as the embossed license plates are metal plates with lettering and numbering raised, unlike the handwritten number plates, and it is fixed onto the vehicle using a `one-way' screw, preventing vehicle owners from changing the license plates on their own. So, the embossed license plates are presented to be tamper-proof and highly secure. That is why an exhaustive geometric mask-based adversarial attack is proposed in this paper. This method of attack was found to be highly probable in a real-life scenario and of high threat to License Plate Recognition Models.

To compare our method to other standard attack methods, Nepal's Embossed License Plate Characters were subjected to gradient-based adversarial attacks such as FGSM, C\&W, and DeepFool to investigate why such gradient-based adversarial images are not practical. After this, the proposed exhaustive geometric mask-based adversarial attack, simulating the physical domain attack, was implemented to generate adversarial images (I-Hard-1057 dataset) of the characters segmented from Nepal's Embossed License Plate (I-1057 dataset). The proposed attack used different geometrical masks for the generation of adversarial images. For this study, three types of patches were used: Horizontal line, Vertical line, and Circular patches. The algorithm to produce adversarial images using Exhaustive Geometric Mask-based Adversarial Attack using Horizontal patches can be seen in Algorithm \ref{alg: exhaustive}.

The proposed method of attack is a white-box attack, as the architecture and structure of the targeted model are accessible, and the model loss is used as a reference to determine the suitable adversarial images. Taking Algorithm \ref{alg: exhaustive} as a reference, vertical and circular patches were implemented as well. In order to generate adversarial images by attacking our LPCR model using exhaustive geometric mask-based adversarial attack, we perturbed benign images from the I-1057 dataset and passed the perturbed image to our model where several parameters are observed: prediction label, confidence of prediction, and loss. The perturbation is performed in an incremental approach where we start from the least noise and slowly increase the amount of perturbation until the LPCR model misclassifies the image with high confidence or the noise reaches its threshold. This process was iterated for every benign image for horizontal line, vertical line, and circular geometric patches.

For a step-by-step algorithm to generate adversarial samples using exhaustive geometric patch-based adversarial attack with minimal mean square error and maximum loss, aiming to achieve misclassification, we begin by initializing several variables. The input variable, denoted as $X$, represents the original image, while the output variable, denoted as $X'$, represents the perturbed image. We also introduce a boolean variable, ``success," which indicates whether the image was misclassified by the model. Additionally, we maintain a counter, ``hit," to keep track of the number of misclassifications.

To control the perturbation level, we introduce the variable ``thickness," representing the amount of perturbation applied to a horizontal patch within the image. The variables $x$ and $y$ represent the width and height of the image, respectively. Lastly, ``rgb" represents the darkest pixel value within the image.

To simulate a realistic attack, we set a threshold for the amount of perturbation. For horizontal patches, we use a threshold equal to half the height of the image, i.e., $y/2$. Starting with an initial thickness of 1, we iterate through thickness values up to half the height of the image, i.e., $y/2$.

For each thickness value, we initialize a variable, ``i," representing the position of the perturbation. It ranges from 0 to $y-$thickness$+1$, which corresponds to the last possible position in the image. We generate a perturbed image, $X'$, by applying the function $perturbimage$, which takes the original image $X$, the perturbation position $i$, and the thickness of the perturbation as parameters. The $perturbimage$ function applies a geometric mask to the benign images based on the provided parameters, generating perturbed images.

Next, we evaluate whether the perturbed image satisfies the necessary conditions to proceed further. If the original image ($X$) is not equal to the perturbed image ($X'$) generated by the $perturbimage$ function, and the prediction loss is greater than the previously recorded loss, this can be considered a better adversarial sample with the maximum loss for the given thickness value. The perturbation parameters are temporarily stored for reference. Specifically, we save the current position as $y'$, the loss as $loss'$, and increment the ``hit" counter by 1.

Once we have applied the perturbation to all possible positions in the image for the given thickness, we analyze the value of ``hit" to determine if we have successfully perturbed the image and selected the one with the maximum loss. If the value of ``hit" is non-zero, indicating at least one misclassification occurred, we use the temporarily stored perturbation parameters to create the best-perturbed image generated by the algorithm with the minimum thickness and maximum loss. In this case, we set the ``success" variable to true.

However, if ``hit" is zero, meaning we were unable to generate a perturbed image that could successfully misclassify the LPCR modal, we increase the thickness and continue the process until we find a perturbed image with the minimum perturbation and maximum loss within that thickness range.

Finally, if ``success" is true, we return the perturbed image as the result; otherwise, we return the original image.

\subsection{External Validation}
It is essential to validate the produced result externally. For this paper, EasyOCR is used to test the generated adversarial images. EasyOCR is a well-known character recognition open-source tool that uses various recognition models such as ResNet, LSTM, and CTC, making it ideal for external validation. During the external validation, all the generated adversarial samples from our model were given as input, and the predicted label and its confidence were noted. The noted result was used to verify if the transferability property holds true (see Supplementary \textbf{Table S2}).

\subsection{Adversarial Training}
Adversarial training is presented as a defense mechanism for the proposed exhaustive geometric mask adversarial attack, where adversarial samples generated using random adversarial parameters are also included in the training dataset (I-Adversarial-1057 dataset). Since the model was subjected to an adversarial attack, evidence to support the existing vulnerability of Nepal's Embossed License Plate Recognition model was found. The study found severe threats to the Embossed License Plate that individuals can use for personal benefit at the expense of breaching the law. This reason leads to the implementation of adversarial training as a defense mechanism. The CNN model was subjected to Adversarial Training with 50\% original and 50\% perturbed dataset. The perturbed dataset consisted of adversarial samples generated using mutually exclusive horizontal line, vertical line, and circular patches with random positions and dimensions during the model's training. This method proved highly effective as the misclassification was significantly reduced.

\begin{algorithm}
    \caption{Exhaustive geometric mask-based adversarial attack using horizontal patches}
    \label{alg: exhaustive}
    \begin{algorithmic}[1]
        \Require $X$
        \Ensure $X'$
        \State $success \gets False$
        \State $hit \gets 0$
        \State $loss' \gets 0$
        \State $thickness \gets 1$
        \State $x \gets width$
        \State $y \gets height$
        \State $rgb[3] \gets darkestpixel[3]$
        \While{$thickness \leq \frac{y}{2}$}
            \For{$i \gets 0$ to ($y-thickness+1$)}
                \State {$X' \gets perturbimage(X, i, thickness)$}\Comment{Perturb $X$ at position $i$}
                \If{$X' \neq X $}
                    \If{$loss > loss'$}\Comment{Select $X'$ with maximum loss}
                        \State{$y' \gets i$}
                        \State{$loss' \gets loss$}
                        \State{$hit \gets hit+1$}
                    \EndIf
                \EndIf
            \EndFor
            \If{$hit \neq 0$}
                \State{$X' \gets perturbimage(X, y', thickness)$}
                \State{$success \gets True$}
                \State break
            \EndIf
            \State $thickness \gets thickness + 1$
        \EndWhile
        \If{$success = True$}
            \State \Return {$X'$}\Comment{Successful to perturb $X$}
        \ElsIf{$success = False$}
            \State \Return {$X$}\Comment{Unsuccessful to perturb $X$}
        \EndIf
    \end{algorithmic}
\end{algorithm}

\section{Results and Analysis}

\subsection{Accuracy of the License Plate Character Recognition (LPCR) model}

Our License Plate Character Recognition (LPCR) model was trained on a dataset consisting of 1057 character images (I-1057 dataset), which were subjected to rotation and blur augmentation while preserving the RGB color dimension. The dataset was divided into an 80-20 train-validation split to assess the model's performance. During training, the model was fine-tuned with specific parameter settings: a Learning Rate of 0.001, a Momentum of 0.9, a Batch size of 64, and Epochs set to 80. This training configuration resulted in the model achieving a validation accuracy of 99.5\%.

\subsection{Adversarial Example Generation}

\begin{table*}[ht]
  \centering
  \caption{Performance of our LPCR and AA-LPCR model demonstrated via 12 example cases. Sample images selected were classified by LPCR (second column) and then perturbed to obtain an adversarial image (third and fourth columns). These adversarial images were reclassified using LPCR (second-last column) and AA-LPCR (last column).}
  \label{tab:attackOutput}
  \scriptsize
  \begin{tabular}{cccccc}\toprule
  \multirow{2}{*}{Original Image} & LPCR Predicted Label & \multirow{2}{*}{Image Perturbation Algorithm} & \multirow{2}{*}{Adversarial Image} & LPCR Predicted Label & Attack-aware LPCR\\
  & \textconfidence(Confidence) & & & \textconfidence(Confidence) & \textconfidence(Confidence)\\\midrule

  % FGSM
    \begin{minipage}{\highlightsize\textwidth}
        \includegraphics[width=\linewidth]{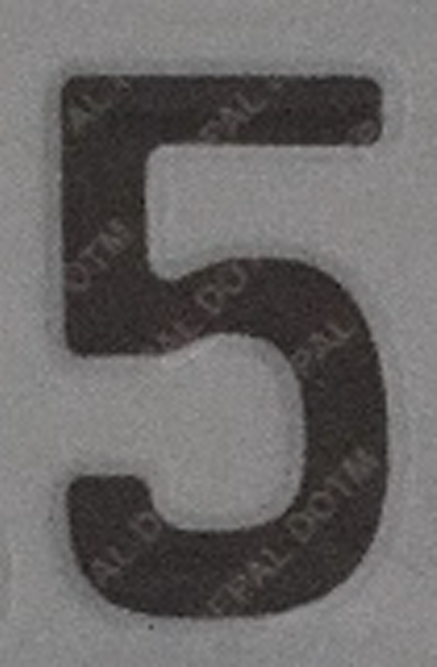}\vspace{1.1pt}
    \end{minipage} & 5 \textconfidence(99.9\%) & \multirow{4}{*}{Fast Gradient Sign Method (FGSM)} & 
    \begin{minipage}{\highlightsize\textwidth}
      \includegraphics[width=\linewidth]{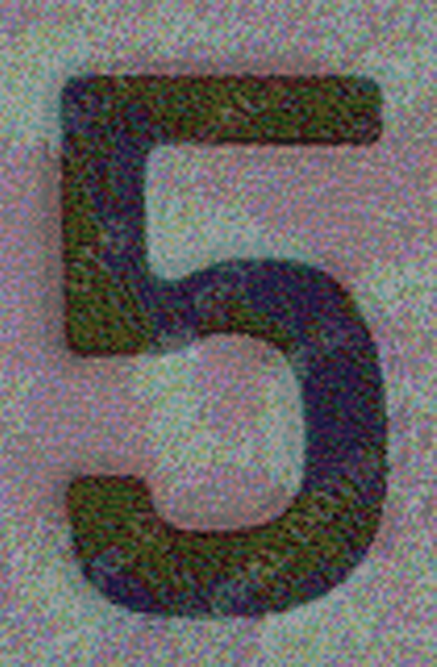}\vspace{1.1pt}
    \end{minipage} & B \textconfidence(68.9\%) & 5 \textconfidence(100.0\%) \\\cmidrule{1-2}\cmidrule{4-6}
    
    \begin{minipage}{\highlightsize\textwidth}
      \includegraphics[width=\linewidth]{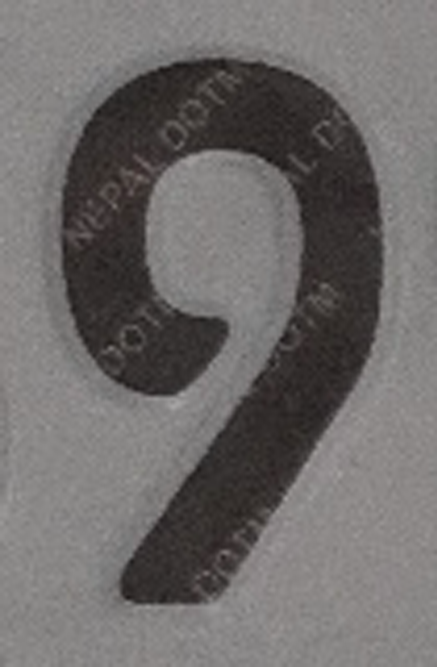}\vspace{1.1pt}
    \end{minipage} & 9 \textconfidence(99.9\%) && 
    \begin{minipage}{\highlightsize\textwidth}
      \includegraphics[width=\linewidth]{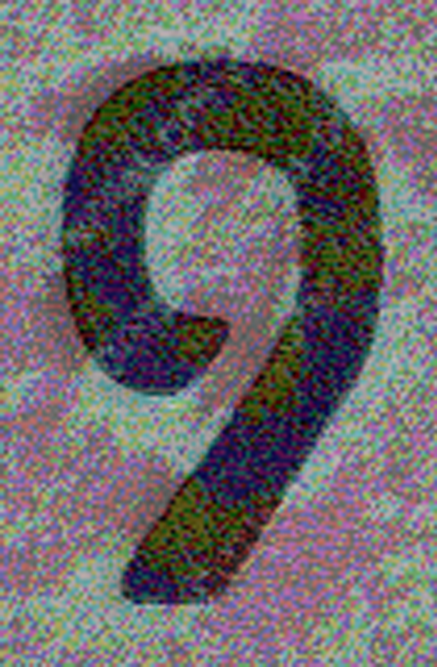}\vspace{1.1pt}
    \end{minipage} & B \textconfidence(96.5\%) & 9 \textconfidence(100.0\%) \\\midrule

     % C&W
    \begin{minipage}{\highlightsize\textwidth}
      \includegraphics[width=\linewidth]{5.png}\vspace{1.1pt}
    \end{minipage} & 5 \textconfidence(99.9\%) & \multirow{4}{*}{Carlini and Wagner attack (C\&W)} & 
    \begin{minipage}{\highlightsize\textwidth}
      \includegraphics[width=\linewidth]{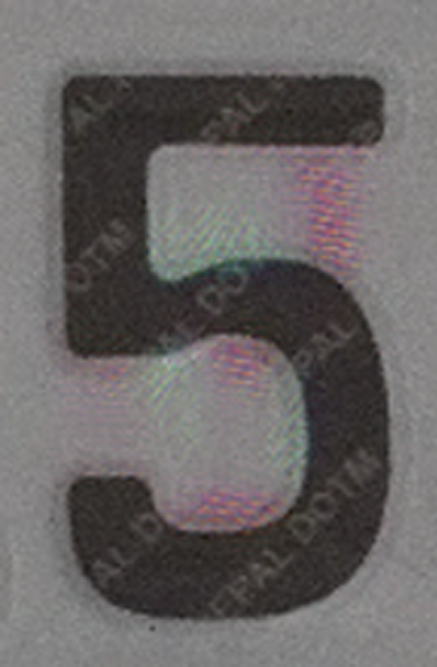}\vspace{1.1pt}
    \end{minipage} & B \textconfidence(52.2\%) & 5 \textconfidence(100.0\%) \\\cmidrule{1-2}\cmidrule{4-6}
    
  \begin{minipage}{\highlightsize\textwidth}
      \includegraphics[width=\linewidth]{9.png}\vspace{1.1pt}
    \end{minipage} & 9 \textconfidence(99.9\%) && 
    \begin{minipage}{\highlightsize\textwidth}
      \includegraphics[width=\linewidth]{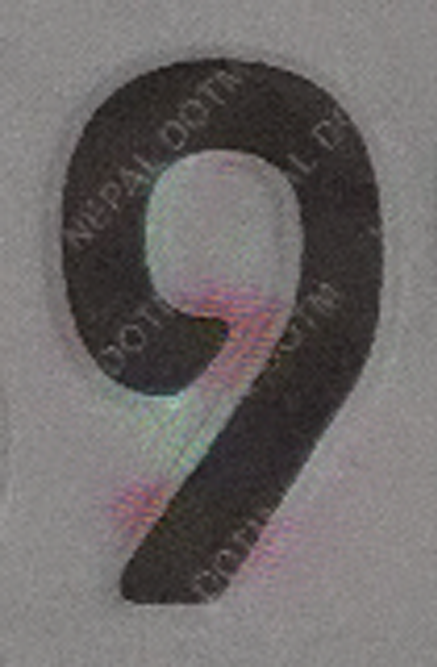}\vspace{1.1pt}
    \end{minipage} & B \textconfidence(50.6\%) & 9 \textconfidence(100.0\%) \\\midrule

     % DeepFool
    \begin{minipage}{\highlightsize\textwidth}
      \includegraphics[width=\linewidth]{5.png}\vspace{1.1pt}
    \end{minipage} & 5 \textconfidence(99.9\%) & \multirow{4}{*}{DeepFool} & 
    \begin{minipage}{\highlightsize\textwidth}
      \includegraphics[width=\linewidth]{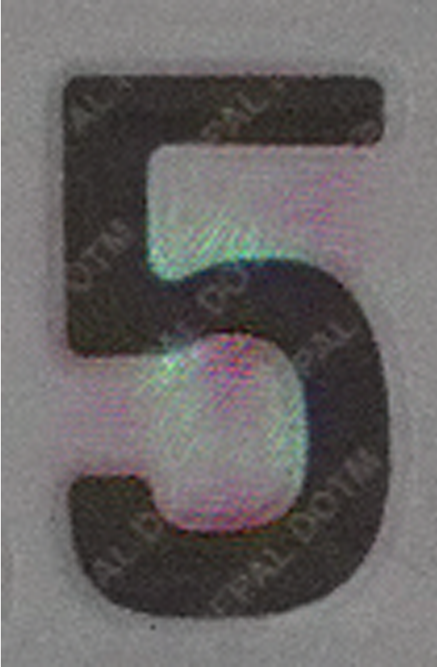}\vspace{1.1pt}
    \end{minipage} & B \textconfidence(53.6\%) & 5 \textconfidence(100.0\%) \\\cmidrule{1-2}\cmidrule{4-6}
    
  \begin{minipage}{\highlightsize\textwidth}
      \includegraphics[width=\linewidth]{9.png}\vspace{1.1pt}
    \end{minipage} & 9 \textconfidence(99.9\%) && 
    \begin{minipage}{\highlightsize\textwidth}
      \includegraphics[width=\linewidth]{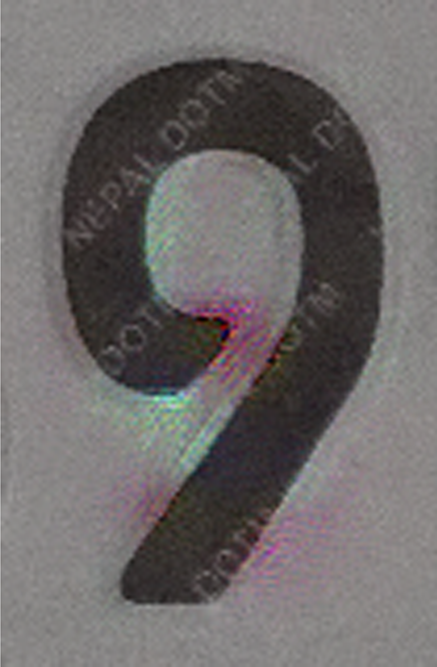}\vspace{1.1pt}
    \end{minipage} & B \textconfidence(50.2\%) & 9 \textconfidence(100.0\%) \\\midrule

  % HLine
    \begin{minipage}{\highlightsize\textwidth}
      \includegraphics[width=\linewidth]{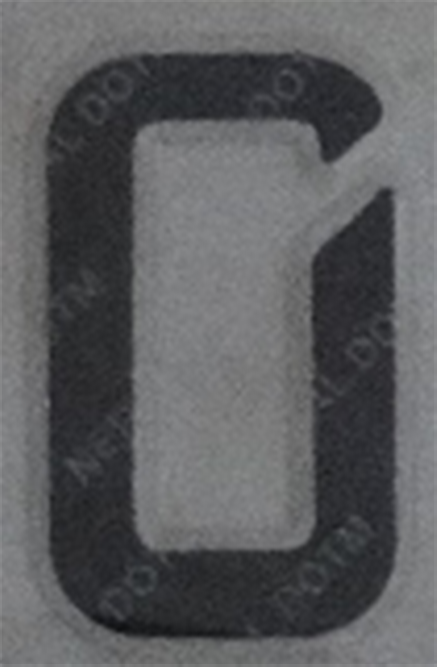}\vspace{1.1pt}
    \end{minipage} & 0 \textconfidence(99.9\%) & \multirow{4}{*}{\shortstack[c]{Adding horizontal lines with\\calculated thickness (this work)}} & 
    \begin{minipage}{\highlightsize\textwidth}
      \includegraphics[width=\linewidth]{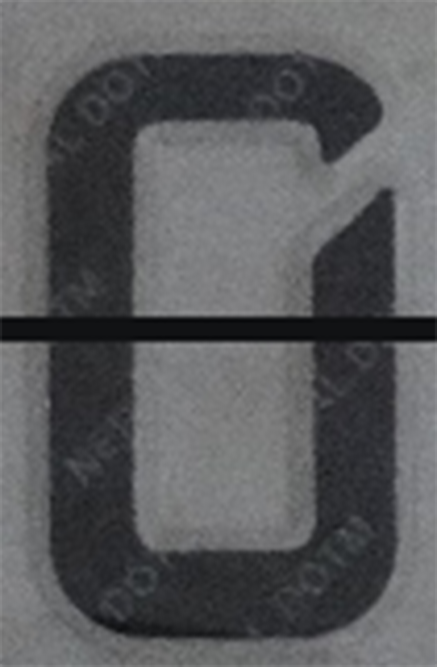}\vspace{1.1pt}
    \end{minipage} & B \textconfidence(72.8\%) & 0 \textconfidence(100.0\%) \\\cmidrule{1-2}\cmidrule{4-6}
    
  \begin{minipage}{\highlightsize\textwidth}
      \includegraphics[width=\linewidth]{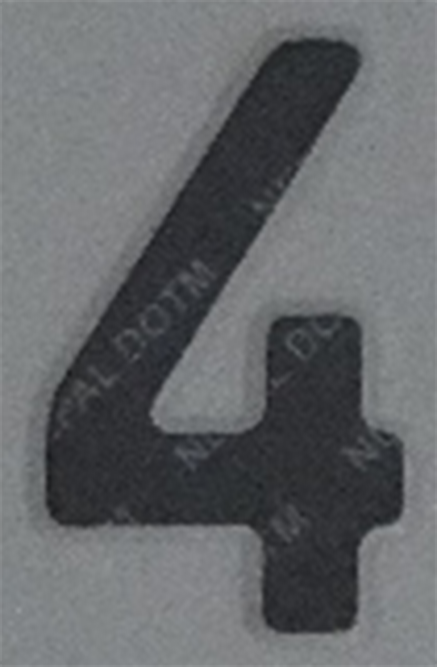}\vspace{1.1pt}
    \end{minipage} & 4 \textconfidence(99.9\%) && 
    \begin{minipage}{\highlightsize\textwidth}
      \includegraphics[width=\linewidth]{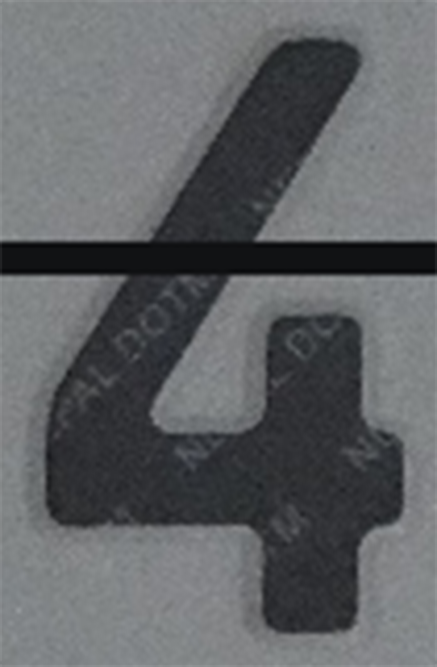}\vspace{1.1pt}
    \end{minipage} & A \textconfidence(58.3\%) & 4 \textconfidence(100.0\%) \\\midrule

     % VLine
    \begin{minipage}{\highlightsize\textwidth}
     \includegraphics[width=\linewidth]{0.png}\vspace{1.1pt}
    \end{minipage} & 0 \textconfidence(99.9\%) & \multirow{4}{*}{\shortstack[c]{Adding vertical lines with\\calculated thickness (this work)}} & 
    \begin{minipage}{\highlightsize\textwidth}
      \includegraphics[width=\linewidth]{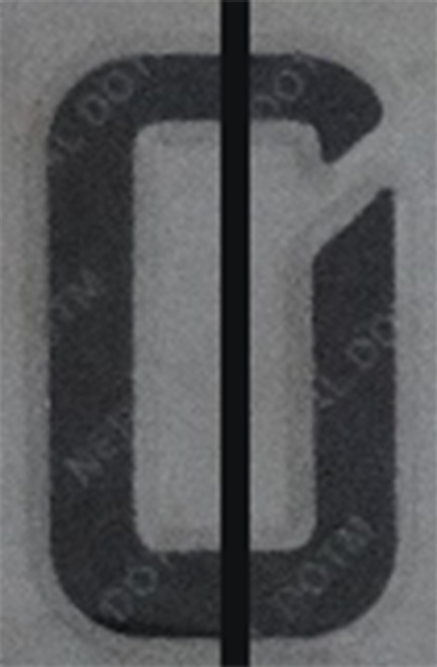}\vspace{1.1pt}
    \end{minipage} & B \textconfidence(75.5\%) & 0 \textconfidence(100.0\%) \\\cmidrule{1-2}\cmidrule{4-6}
    
  \begin{minipage}{\highlightsize\textwidth}
      \includegraphics[width=\linewidth]{4.png}\vspace{1.1pt}
    \end{minipage} & 4 \textconfidence(99.9\%) && 
    \begin{minipage}{\highlightsize\textwidth}
      \includegraphics[width=\linewidth]{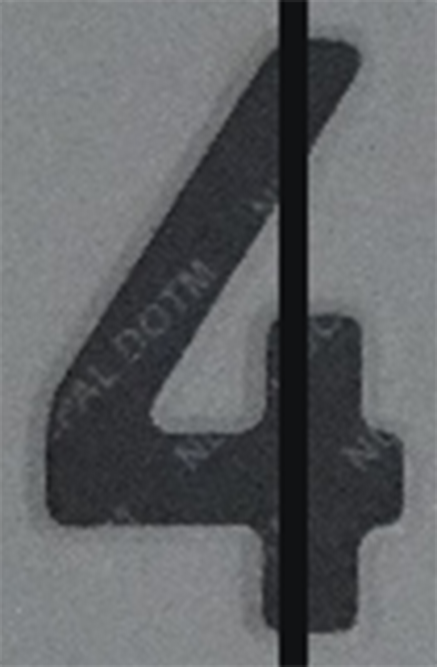}\vspace{1.1pt}
    \end{minipage} & A \textconfidence(67.8\%) & 4 \textconfidence(100.0\%) \\\midrule

     % Spot
    \begin{minipage}{\highlightsize\textwidth}
      \includegraphics[width=\linewidth]{0.png}\vspace{1.1pt}
    \end{minipage} & 0 \textconfidence(99.9\%) & \multirow{4}{*}{\shortstack[c]{Adding circular spots with \\calculated radius (this work)}} & 
    \begin{minipage}{\highlightsize\textwidth}
      \includegraphics[width=\linewidth]{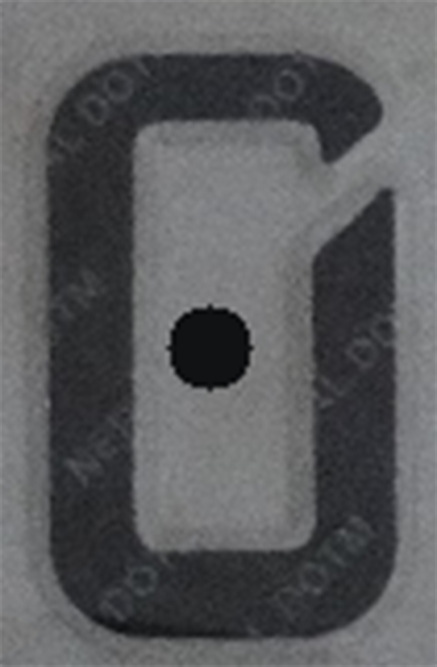}\vspace{1.1pt}
    \end{minipage} & B \textconfidence(84.9\%) & 0 \textconfidence(100.0\%) \\\cmidrule{1-2}\cmidrule{4-6}
    
  \begin{minipage}{\highlightsize\textwidth}
      \includegraphics[width=\linewidth]{4.png}\vspace{1.1pt}
    \end{minipage} & 4 \textconfidence(99.9\%) && 
    \begin{minipage}{\highlightsize\textwidth}
      \includegraphics[width=\linewidth]{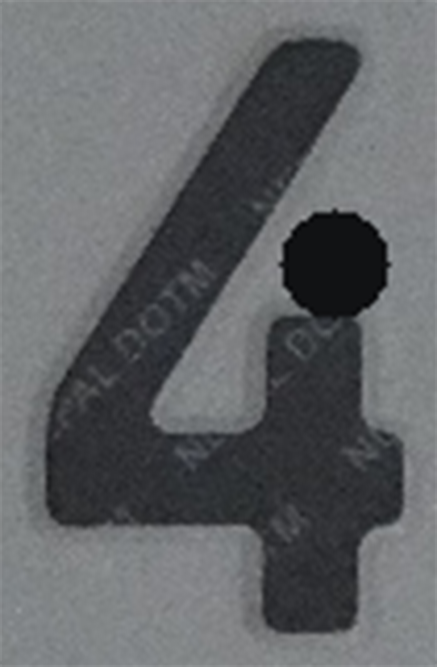}\vspace{1.1pt}
    \end{minipage} & A \textconfidence(78.3\%) & 4 \textconfidence(100.0\%) \\
    \bottomrule
  \end{tabular}
\end{table*}

Following the standard practice of generating adversarial samples by perturbing an image until a model misclassifies the image, we tried several methods to generate adversarial samples for our I-1057 dataset. While commonly used Gradient-based Adversarial Attacks such as FGSM \cite{goodfellow2014explaining}, C\&W \cite{carlini2017towards}, and DeepFool \cite{moosavi2016deepfool} are readily available to use, we observed that the samples generated by these methods contain specific color taints throughout the image that appear unrealistic. For a common real-world attack, intentional or inadvertent, changing an entire license plate in a specific way may not be practical. Some examples of these samples are shown in the Adversarial Sample column in \textbf{Table \ref{tab:attackOutput}}.

To simulate more realistic attacks, we performed a white-box attack on the LPCR model using our exhaustive geometric mask algorithm. In order to generate adversarial images by attacking our LPCR model, we perturbed I-1057 images and passed the perturbed image to our model, then observed several parameters: prediction label, confidence of prediction, and loss. The perturbation is performed in an incremental approach where we start from the least noise and slowly increase the amount of perturbation until the LPCR model misclassifies the image with high confidence and minimum noise or the noise reaches its threshold. This process was iterated for every benign image for different geometric patches. Samples of generated adversarial images can be seen in \textbf{Table \ref{tab:attackOutput}}.

Next, adversarial samples were generated by attacking the LPCR model using our exhaustive geometric mask algorithm (see Methods). Using the original 1057 character images (I-1057 dataset), several random adversarial images were generated to test the performance of LPCR. A significant drop in the performance of the LPCR model was observed on the generated random adversarial images. Exhaustively perturbing I-1057 images to attack the LPCR model, we identified all the “hard” variants of I-1057 images and grouped them as the I-Hard-1057 dataset. Our LPCR model correctly classified only 24\% of images in this I-Hard-1057 dataset. Most of the correctly classified images were from class `A', i.e., `A' was least vulnerable in comparison to other characters (see Supplementary \textbf{Table S1}). Overall, this drastic drop in the accuracy of the LPCR model validates that a highly accurate model is not necessarily robust. 

We also studied the amount of change (in pixels) needed in the images for them to be misclassified. To analyze this, we calculate the Mean Squared Error (MSE) between the original image and its perturbed variant. MSE is used to quantify the amount of perturbation added by measuring the average distance between the original image and the perturbed image. The analysis presented in \textbf{Table \ref{tab:comparison on the basis of characters}} reveals notable distinctions in the MSE among adversarial samples generated using vertical, horizontal, and circular patches. Adversarial samples created with vertical patches exhibit the highest MSE on average, signifying their conspicuous nature. In contrast, adversarial samples generated with circular patches show the lowest MSE, implying a more unobtrusive distortion. Furthermore, circular patches yield the highest confidence level, reinforcing their efficacy. When evaluated holistically, the data indicates that circular patches provide the most effective and efficient performance among the three patch shapes, given their ability to generate adversarial samples with minimal deviation from the original samples, and with a high degree of confidence. Therefore, circular patches emerge as the most optimal choice for generating inconspicuous adversarial samples.

\begin{table*}[ht]
  \centering
  \caption{Average metrics showing the effectiveness of generated adversarial samples against LPCR using average confidence of misclassification and average mean squared error (MSE). The average MSE represents the average squared difference between the original and adversarial images (amount of perturbation). The higher value of average confidence shows a higher ability of the adversarial image to fool the LPCR model and vice versa. The ideal condition is to have higher confidence of misclassification and lower MSE from an attacker's perspective and vice versa from a defender's perspective.}
  \label{tab:comparison on the basis of characters}
  \scriptsize
  \begin{tabular}{crrrrrrr}\toprule
  \multirow{2}{*}{Characters} &\multicolumn{2}{c}{Horizontal Patches} &\multicolumn{2}{c}{Vertical Patches} &\multicolumn{2}{c}{Circular Patches} \\\cmidrule{2-7}
  & Avg. Confidence &Avg. MSE & Avg. Confidence &Avg. MSE & Avg. Confidence &Avg. MSE \\\midrule
  0 &69.3 &366 &79.6 &647 &77.5 &294 \\
  1 &51.3 &1551 &48.2 &1292 &56.2 &1085 \\
  2 &57.2 &756 &61.6 &658 &65.2 &689 \\
  3 &59.2 &460 &57.6 &336 &63.3 &397 \\
  4 &53.6 &562 &68.0 &499 &78.0 &326 \\
  5 &60.0 &547 &60.2 &578 &75.2 &279 \\
  6 &57.0 &596 &61.6 &545 &62.0 &395 \\
  7 &57.6 &447 &58.6 &947 &71.0 &402 \\
  8 &54.5 &1985 &53.6 &657 &62.5 &906 \\
  9 &53.1 &781 &45.2 &1646 &59.6 &605 \\
  A &47.7 &1062 &47.3 &1212 &68.7 &970 \\
  B &64.5 &459 &51.3 &852 &71.5 &413 \\
  F &51.8 &818 &54.4 &1090 &73.6 &553 \\
  \midrule
  Average &56.7 &799 &57.5 &859 &68.0 &562 \\
  \bottomrule
  \end{tabular}
\end{table*}

\begin{figure*}
\centering
  \subfloat[]{%
    \includegraphics[width=\heatmapsize\textwidth]{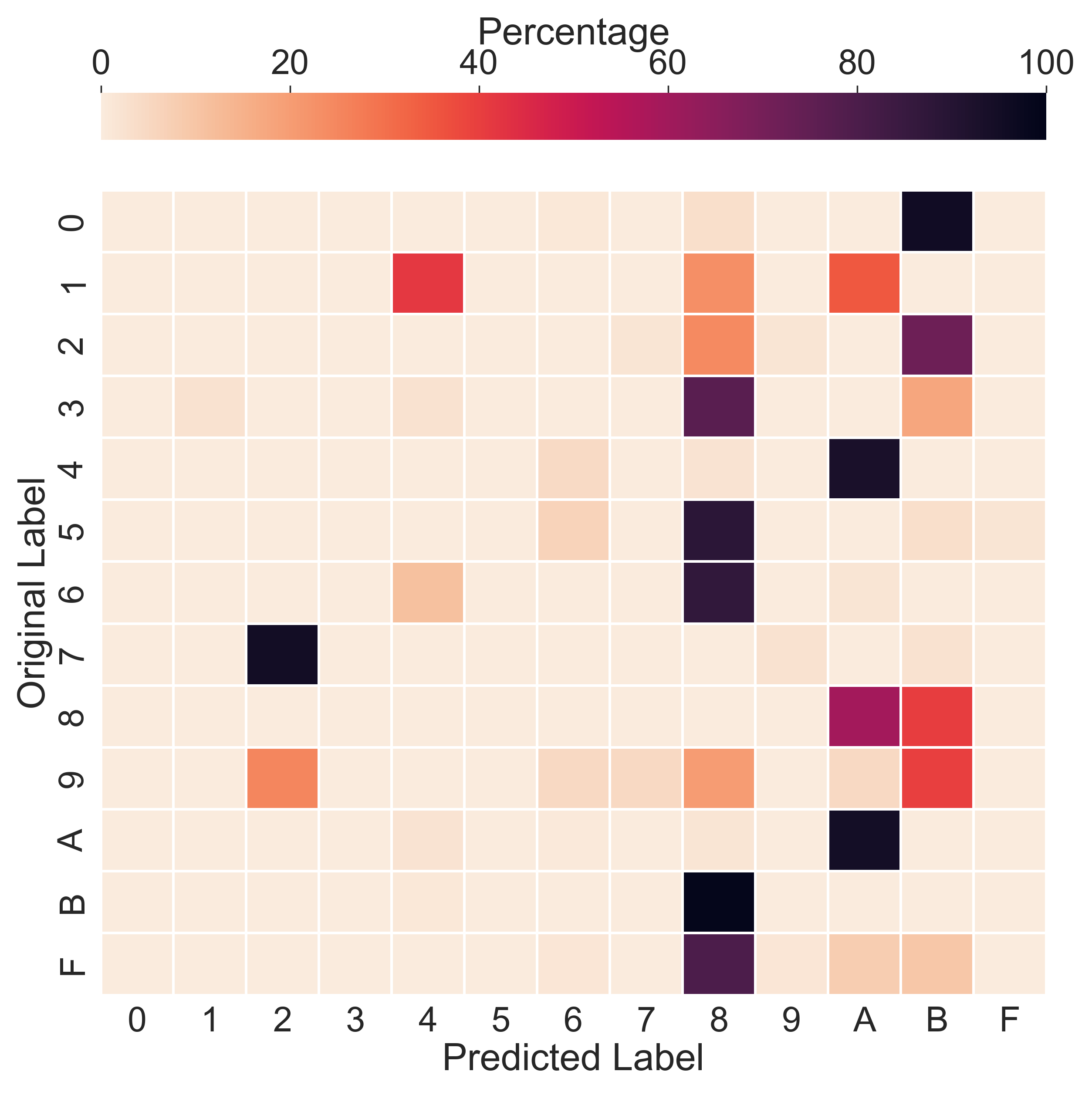}}\hspace{0.1em}%\hfill
  \subfloat[]{%
    \includegraphics[width=\heatmapsize\textwidth]{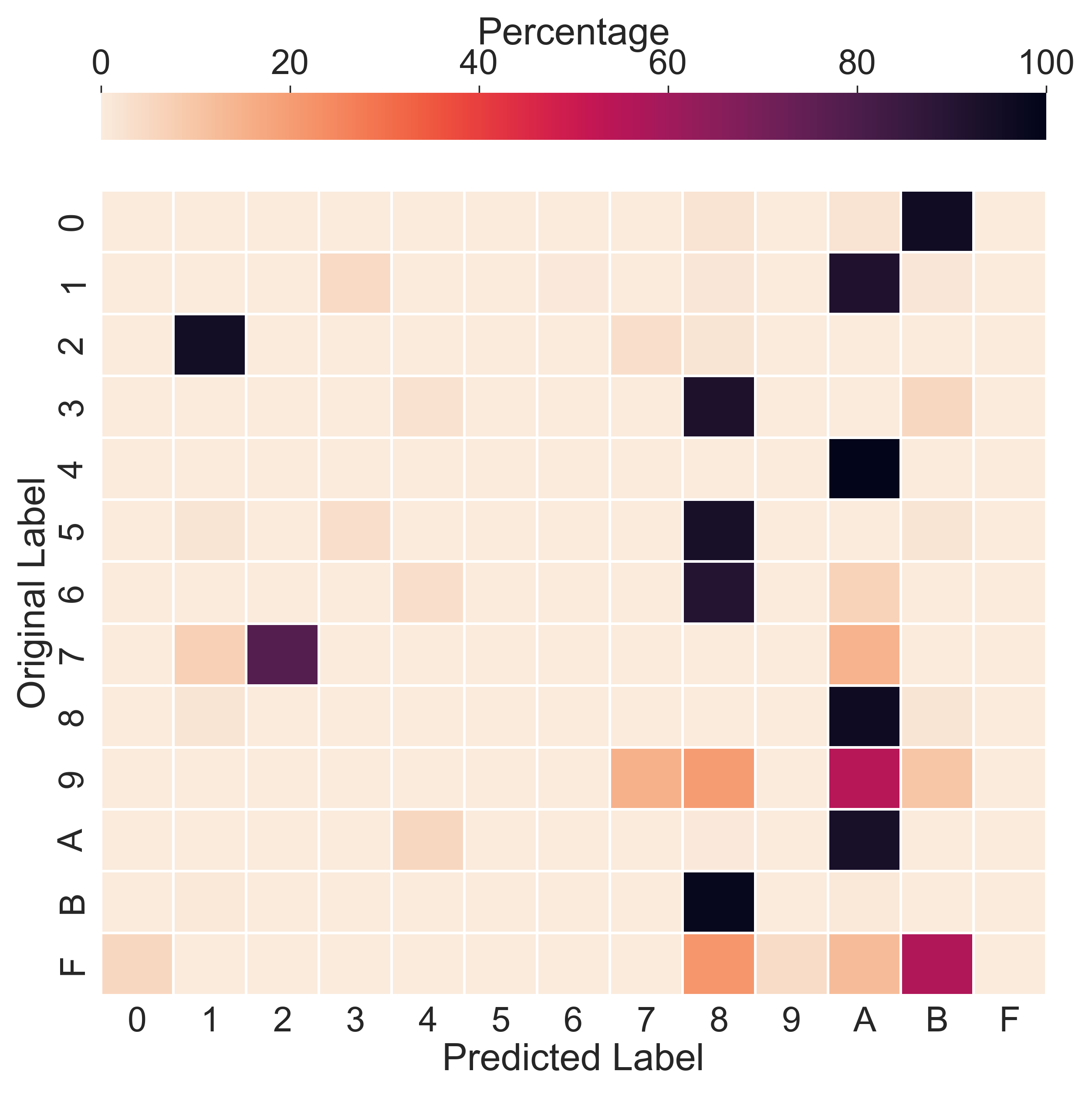}}\hspace{0.1em}%\hfill
  \subfloat[]{%
    \includegraphics[width=\heatmapsize\textwidth]{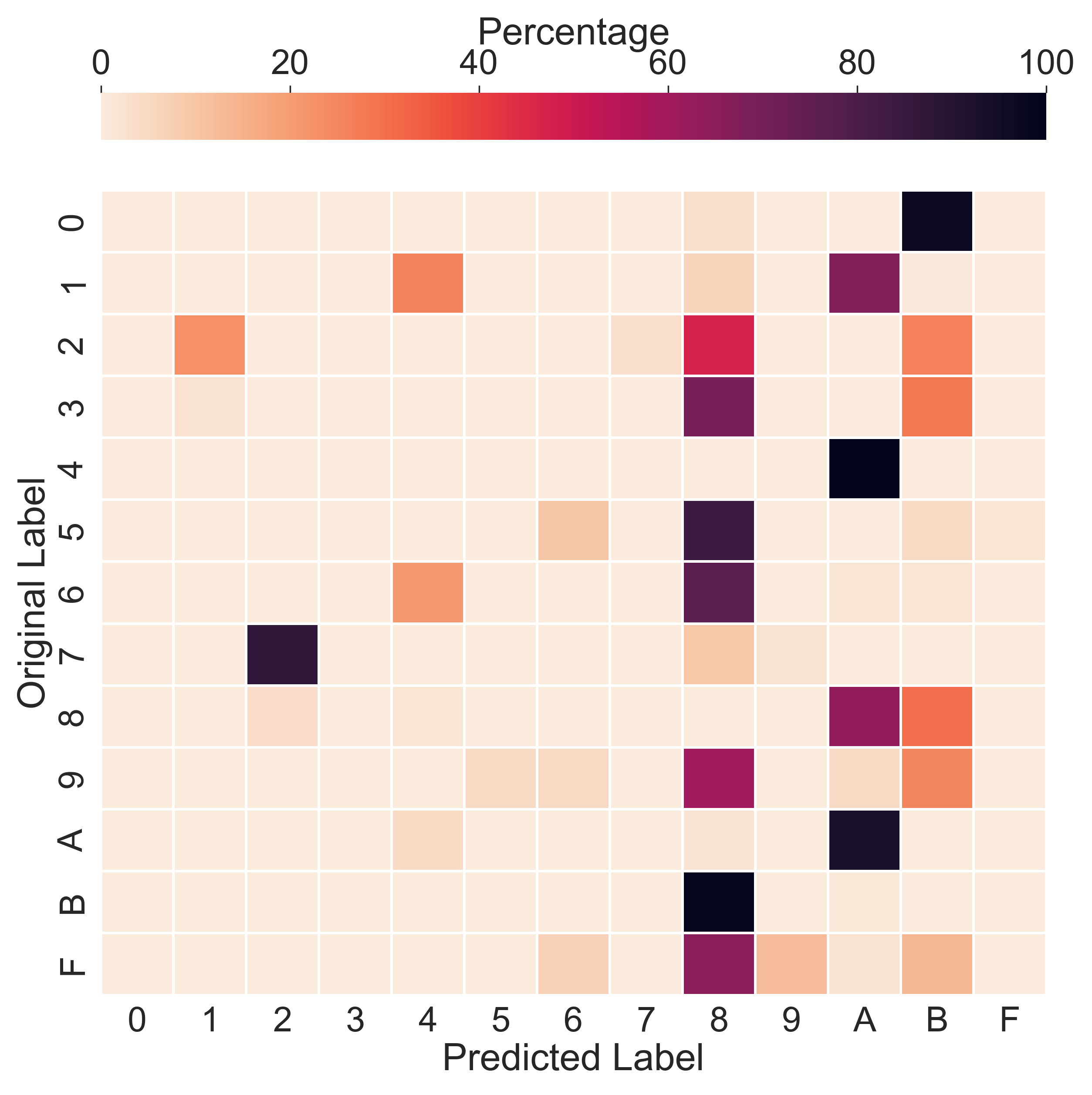}}
  \caption{The heatmap shows the percentage of images that were classified correctly (cells in the diagonal) and incorrectly (cells not in the diagonal) when adversarial patches are added to the images. The patches applied are horizontal line \textbf{(a)}, vertical lines \textbf{(b)}, and circular \textbf{(c)}. For instance, in \textbf{(a)}, the cell in the first row and second-last column implies that the label `0' was heavily misclassified as `B'. Cells in the left-to-right diagonal line represent correct classifications. From the heatmaps, it can be observed that the exhaustive geometric mask-based adversarial attack was able to misclassify most of the characters efficiently apart from the character `A' regardless of whether the patches are horizontal \textbf{(a)}, vertical \textbf{(b)}, or circular \textbf{(c)}.
  }
  \label{fig:heatmap}
\end{figure*}

\begin{figure*}
\centering
    \includegraphics[width=\textwidth]{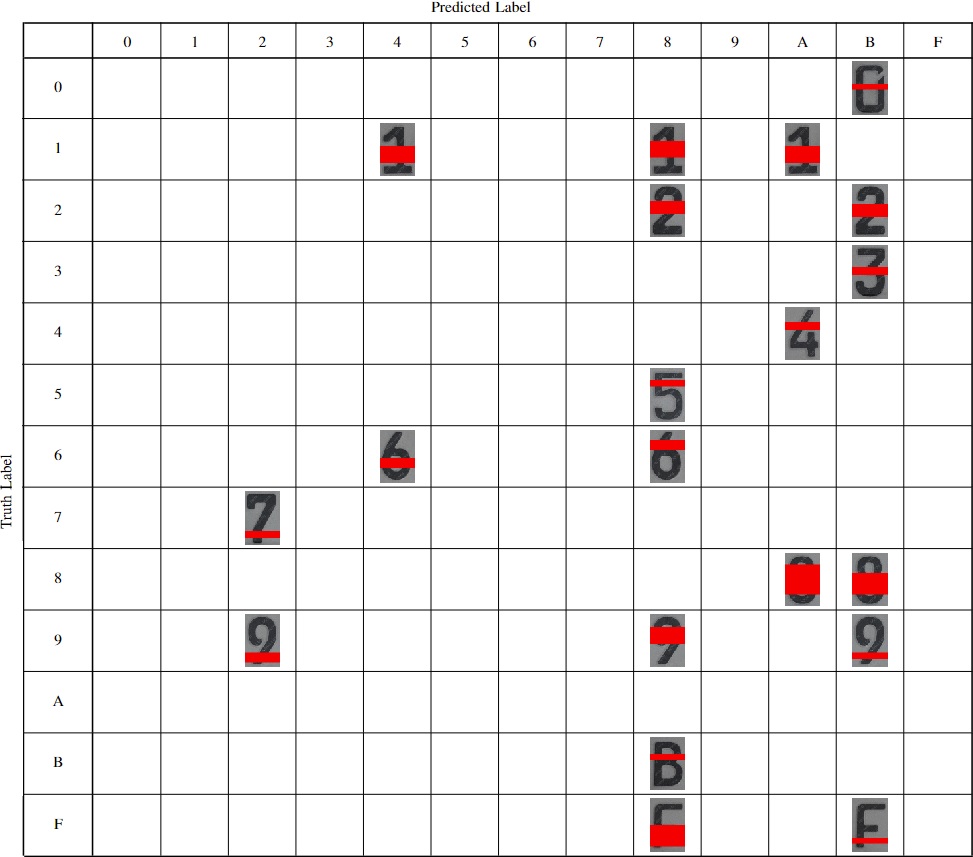}
    \caption{Attack-prone regions of license plate character images identified by predicting the class of the images after adding horizontal line patches. For instance, images of a character `0' (row `0') are highly likely to be misclassified as a `B' (column `B') if adversarial horizontal line patches appear in the red highlighted region. Perturbed images were passed through our license plate character recognition (LPCR) model to predict their labels and were analyzed to obtain these highlighted regions.}
    \label{fig:attack-prone-regions}
\end{figure*}

Analyzing the adversarial image samples, we found some common patterns. These patterns were based either on the class of the image (i.e., what character it is) or the location and direction of the patch introduced. Analyzing these patterns, we were able to find vulnerable areas in the character images. First, we studied the average probabilities that any given character image may be misclassified as another character. Our goal in this task was to find answers to questions of the form, ``How likely can a `0' be misclassified as another character when we apply a horizontal patch?" Our findings, summarized in \textbf{Figure \ref{fig:heatmap}} show several notable misclassifications. Regardless of the type of patch applied (horizontal, vertical, or circular), we find that most characters are prone to be misclassified as either `8', `A', or `B'. Similarly, the most common misclassifications include `0' misclassified as `B', `4' as `A', and `7' as `2'. In general, across all three patch types, we found that the images of the character `A' were the least misclassified, suggesting that it is the most difficult character to attack. Next, for the case of introducing horizontal patches, we visualized the misclassified images and the vulnerable regions within the images (see \textbf{Figure \ref{fig:attack-prone-regions}}). Visualization of the character images, along with the highlighted vulnerable bands on them, shows that several character images, such as `0' and `5', have smaller regions of vulnerability regions. On the other hand, images such as `8' and `1' have much larger attack-prone regions.

% Heatmap
% \begin{figure*}[htb]

\subsection {Comparison of LPCR with EasyOCR}

Since our LPCR model was susceptible to our exhaustive geometric mask-based adversarial attack (i.e., the attack was successful), we sought to check if the same I-Hard-1057 could be successful at attacking other state-of-the-art character recognition models. To validate the transferability of these results, to classify the images in the I-Hard-1057 dataset we used EasyOCR, a Convolutional Recurrent Neural Network (CRNN) character recognition model \cite{shi2015endtoend}. EasyOCR is a widely-used, open-source, lightweight character recognition tool recognized for its robust text recognition capabilities. Its extensive versatility enables it to excel in diverse OCR applications, including text extraction from natural scenes and documents with support for over 80 languages. On our I-Hard-1057 dataset, EasyOCR's accuracy was 29.2\% (see Supplementary \textbf{Table S2}), which is similar to the accuracy of LPCR. This confirms that the adversarial samples generated by our exhaustive geometric mask-based adversarial attack algorithm can also fool other deep learning models.

\subsection{Adversarial Attack-aware License Plate Character Recognition (AA-LPCR) model}

Retraining our LPCR model on the new I-Adversarial-Train dataset (see Methods), we developed an adversarial attack-aware license plate character recognition (AA-LPCR) model and found it to be considerably resilient. 
This AA-LPCR model was able to correctly classify 99.74\% of the I-Hard-1057 dataset, for which the accuracy of the LPCR model was only 24.06\%. The remaining rare I-Hard-1057 images that were misclassified are illustrated in \textbf{Table \ref{tab:rare cases}}. We then proceeded to attack the AA-LPCR model using the exhaustive geometric mask-based adversarial attack and observed that the success rate of the attack significantly dropped as low as 21.95\% from 75.9\% for vertical patches. As illustrated in \textbf{Table \ref{tab: adversarial attack success rate}}, a similar decline can be seen for both horizontal and circular patches as well. This increase in the accuracy of AA-LPCR on the I-Hard-1057 dataset and the decrease in the success rate of exhaustive geometric mask-based adversarial attack on AA-LPCR validates that Adversarial Training is efficient in increasing the robustness of the model.

\begin{table}[ht]
  \centering
  \caption{This figure shows all the cases in which AA-LPCR fails to correctly classify the adversarial images. This demonstrates the limitation of our resilient AA-LPCR model in rare cases.}
  \label{tab:rare cases}
  \scriptsize
  \begin{tabular}{cccccc}\toprule
  \multirow{2}{*}{Adversarial Image} & Predicted Label & Result of our new resilient model\\
  & \textconfidence(Confidence) & \textconfidence(Confidence)\\\midrule

    \begin{minipage}{\highlightsize\textwidth}
      \includegraphics[width=\linewidth]{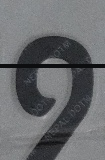}\vspace{1.1pt}
    \end{minipage} & 6 \textconfidence(57.0\%) & 0 \textconfidence(99.9\%) \\\midrule

    \begin{minipage}{\highlightsize\textwidth}
      \includegraphics[width=\linewidth]{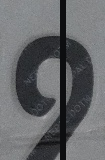}\vspace{1.1pt}
    \end{minipage} & A \textconfidence(54.0\%) & 0 \textconfidence(99.9\%) \\\midrule

    \begin{minipage}{\highlightsize\textwidth}
      \includegraphics[width=\linewidth]{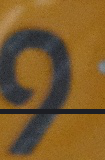}\vspace{1.1pt}
    \end{minipage} & A \textconfidence(57.5\%) & 7 \textconfidence(84.8\%) \\\midrule

    \begin{minipage}{\highlightsize\textwidth}
      \includegraphics[width=\linewidth]{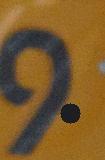}\vspace{1.1pt}
    \end{minipage} & A \textconfidence(74.7\%) & 7 \textconfidence(67.4\%) \\\midrule

    \begin{minipage}{\highlightsize\textwidth}
      \includegraphics[width=\linewidth]{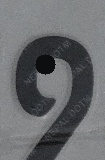}\vspace{1.1pt}
    \end{minipage} & 6 \textconfidence(48.6\%) & 0 \textconfidence(99.9\%) \\\midrule

    \begin{minipage}{\highlightsize\textwidth}
      \includegraphics[width=\linewidth]{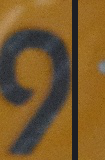}\vspace{1.1pt}
    \end{minipage} & A \textconfidence(76.0\%) & 7 \textconfidence(96.8\%) \\\midrule

    \begin{minipage}{\highlightsize\textwidth}
      \includegraphics[width=\linewidth]{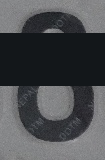}\vspace{1.1pt}
    \end{minipage} & B \textconfidence(52.9\%) & 6 \textconfidence(66.7\%) \\\midrule
    
    \begin{minipage}{\highlightsize\textwidth}
      \includegraphics[width=\linewidth]{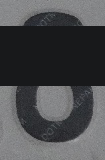}\vspace{1.1pt}
    \end{minipage} & B \textconfidence(49.7\%) & 6 \textconfidence(52.3\%) \\
    \bottomrule
  \end{tabular}
\end{table}

\begin{table}
  \centering
  \caption{Comparison of adversarial attack success rate between License plate character recognition (LPCR) model and Adversarial attack-aware license plate character recognition (AA-LPCR) model.}
  \label{tab: adversarial attack success rate}
  \begin{tabular}{ccc}\toprule
  Geometric Patches & LPCR  & AA-LPCR\\\midrule
  Horizontal & 75.6\% & 56.09\% \\ 
  Vertical & 75.9\% & 21.95\% \\ 
  Circular & 76.1\% & 63.41\% \\ 
  \bottomrule
  \end{tabular}
\end{table}

% === V. CONCLUSION AND FUTURE WORK ========================================
% =================================================================================
\section{Conclusion and Future Work}

Using Nepal's embossed license plate images as a dataset, this work classified license plate characters using standard deep learning models. Findings elucidate that license plate recognition systems designed using off-the-shelf deep learning methods can be easily tricked, intentionally or inadvertently. Overall, we find that existing deep learning-based character recognition methods that are unaware of adversarial attacks can be ineffective for license plate recognition. As a solution, the license plate character recognition method developed herein was refined by retraining using adversarial samples to increase classification accuracy. Developing this adversarial attack-aware license plate character recognition (AA-LPCR) model, we demonstrate that license plate recognition systems can be refined to be remarkably more accurate and effectively be `attack-aware'.

Several additional experiments could improve the prediction accuracy of our AA-LPCR method. In addition to perturbing images by adding vertical, horizontal, and circular patches, several additional geometric shapes, such as squares and rectangles, can be explored to study how they affect prediction accuracy. Also, these shapes can be dulled or blunted and combined to generate more diverse adversarial samples and more effectively simulate real-world physical attacks. Furthermore, a complete attack-aware license plate recognition method can be developed using our attack-aware character recognition models. 
Finally, interpretability experiments can lead to an in-depth understanding of why certain characters and their regions are prone to be misclassified as others.

% === ACKNOWLEDGEMENTS =============================================================
% =================================================================================
\section*{Acknowledgments}
We would like to thank the Vehicle Fitness Test Center (VFTC), Kathmandu, Nepal for allowing us to take pictures of license plates from their inventory.

% === REFERENCE SECTION =============================================================
% =================================================================================
\bibliographystyle{unsrt}
\bibliography{ref}

\vfill

\end{document}